%% file: neurips_2025.tex
\documentclass{article}

\usepackage[final]{neurips_2025}

\usepackage[utf8]{inputenc} % allow utf-8 input
\usepackage[T1]{fontenc}    % use 8-bit T1 fonts
\usepackage{hyperref}       % hyperlinks
\usepackage{url}            % simple URL typesetting
\usepackage{booktabs}       % professional-quality tables
\usepackage{amsfonts}       % blackboard math symbols
\usepackage{nicefrac}       % compact symbols for 1/2, etc.
\usepackage{microtype}      % microtypography
\usepackage{xcolor}         % colors
\usepackage{amsmath}
\usepackage{graphicx}
\usepackage{titletoc}
\usepackage{algorithm}
\usepackage{algorithmic}
\usepackage{multirow}
\title{MetaFind: Scene-Aware 3D Asset Retrieval for Coherent Metaverse Scene Generation}
\usepackage{subcaption}
\usepackage{caption}

\author{%
  Zhenyu Pan\\
  Northwestern University\\
  \texttt{zhenyupan@u.northwestern.edu} \\
  \And
  Yucheng Lu\\
  New York University\\
  \texttt{yuchenglu@nyu.edu}   
  \And
  Han Liu\\
  Northwestern University\\
  \texttt{hanliu@northwestern.edu} \\
}

\begin{document}

\maketitle

\begin{abstract}
We present \textbf{MetaFind}, a scene-aware tri-modal compositional retrieval framework designed to enhance scene generation in the metaverse by retrieving 3D assets from large-scale repositories. MetaFind addresses two core challenges: (i) inconsistent asset retrieval that overlooks spatial, semantic, and stylistic constraints, and (ii) the absence of a standardized retrieval paradigm specifically tailored for 3D asset retrieval, as existing approaches mainly rely on general-purpose 3D shape representation models. Our key innovation is a flexible retrieval mechanism that supports arbitrary combinations of text, image, and 3D modalities as queries, enhancing spatial reasoning and style consistency by jointly modeling object-level features (including appearance) and scene-level layout structures. Methodologically, MetaFind introduces a plug-and-play equivariant layout encoder \textbf{ESSGNN} that captures spatial relationships and object appearance features, ensuring retrieved 3D assets are contextually and stylistically coherent with the existing scene, regardless of coordinate frame transformations. The framework supports iterative scene construction by continuously adapting retrieval results to current scene updates. Empirical evaluations demonstrate the improved spatial and stylistic consistency of MetaFind in various retrieval tasks compared to baseline methods.

\end{abstract}

% \vspace{-0.1in}
\section{Introduction}
\label{sec:introduction}
This work introduces MetaFind, a novel scene-aware 3D retrieval framework designed to facilitate coherent scene generation within the metaverse by retrieving 3D assets from extensive repositories. Effective scene generation heavily relies on retrieving relevant, consistent, and contextually appropriate 3D assets \cite{sun2024layoutvlm}; however, current methods face significant limitations, primarily due to two key challenges. First, existing retrieval frameworks often overlook critical factors such as spatial relationships, semantic coherence, and stylistic consistency, leading to retrieved assets that are visually and contextually incongruous when integrated into complex scenes \cite{liu2023openshape}. Second, unlike well-established retrieval paradigms in natural language processing (NLP), such as Dense Passage Retrieval (DPR) \cite{karpukhin2020dense}—which introduced a generalizable dual-encoder architecture—there is currently no standardized retrieval paradigm explicitly tailored to the requirements and characteristics of 3D asset retrieval. Finally, recent retrieval depends on generic 3D shape representation models, which fail to capture scene-specific contextual and stylistic nuances essential for coherent scene layout. 

\begin{figure*}[t]
    \centering
    \includegraphics[width=1\linewidth]{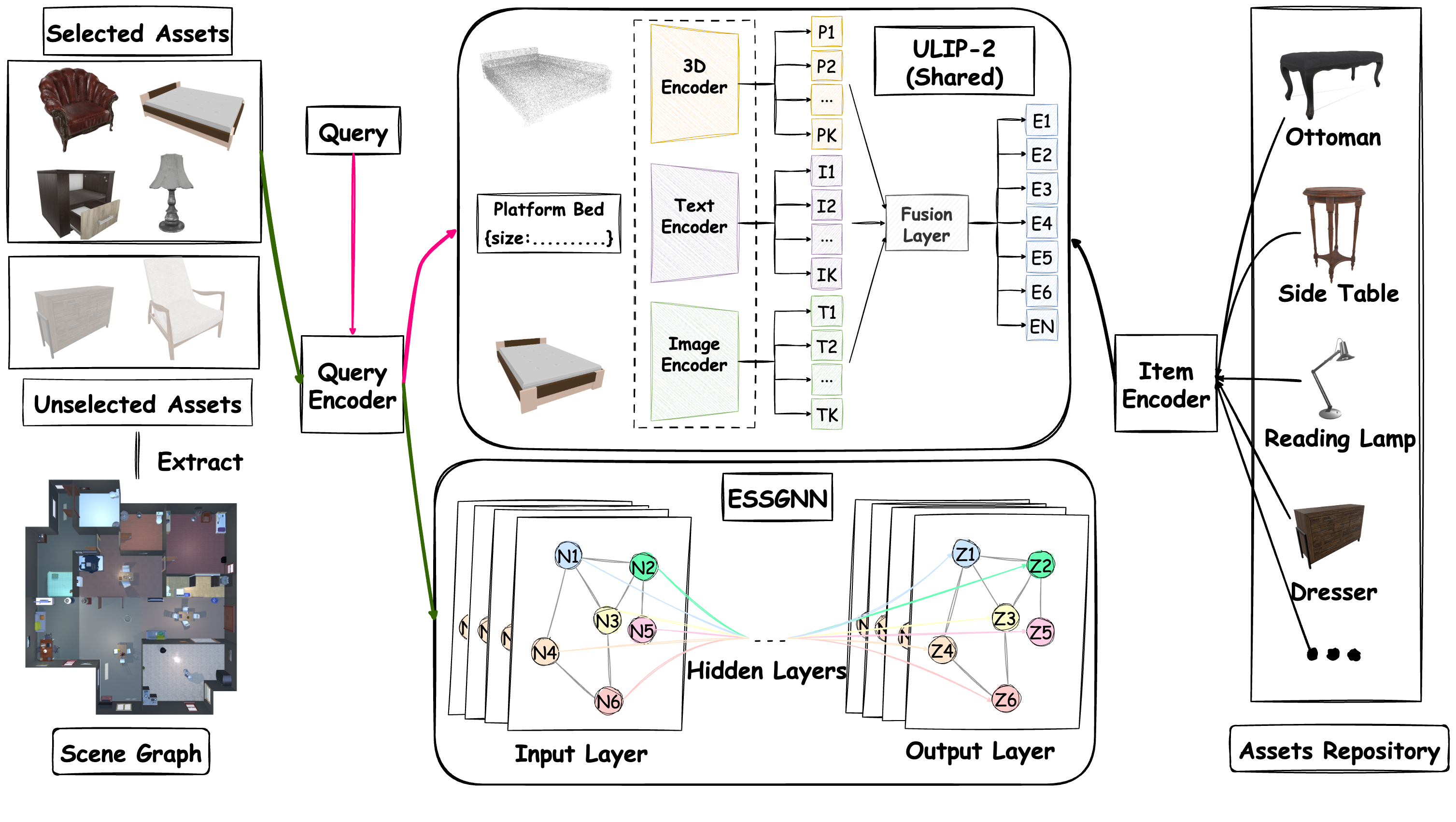}
    % \vspace{-0.3in}
    \caption{Overall framework. MetaFind adopts a dual-tower design where both the user query and candidate assets are encoded using the ULIP-2 backbone. On the query side, we incorporate a plug-and-play ESSGNN module that encodes the current scene layout into a structured scene graph, which captures spatial relationships and object attributes. The user's input—text, image, point cloud, or any combination—is processed by ULIP-2 and fused with the scene context embedding from the ESSGNN to produce a layout-aware query representation. On the asset side, each 3D asset in the repository is pre-encoded independently by ULIP-2 into a fixed vector. At retrieval time, the similarity between the layout-aware query embedding and the precomputed asset embeddings is computed, and the top-matching asset is selected to be inserted into the scene.}
    \label{fig:framework}
    % \vspace{-0.3in}
\end{figure*}

Recent approaches try to address these challenges by introducing various strategies. Early efforts enhance retrieval through 3D representations, focusing on object-level geometric features \cite{guo2015learning,su2015multi}. Subsequent studies address cross-domain retrieval limitations through advanced techniques. Methods like SPL \cite{wang2020unsupervised} leverage domain alignment strategies, minimizing inter-domain discrepancies. UCD \cite{song2020universal} proposes sample-level weighting combined with domain and class alignment mechanisms, achieving improved performance but still relying on labeled data and introducing prediction bias. More recently, S2Mix \cite{fu2025s2mix} and SCA3D \cite{ren2025sca3d} introduce style fusion layers and cross-modal data augmentation techniques to enhance retrieval performance. Despite these improvements, the current approaches are limited as they mainly consider object-centric features without adequately capturing crucial spatial, contextual, and scene-level relationships. Furthermore, they only support single-modality queries (3D-to-3D, text-to-3D, or image-to-3D), lacking the flexibility to handle compositional queries across multiple modalities. To address these limitations, MetaFind introduces a retrieval paradigm that supports compositional multi-modal queries and incorporates spatial reasoning, semantic coherence, and stylistic consistency to ensure seamless integration of retrieved 3D assets into complex scenes.

To this end, we propose MetaFind, a dual-tower retrieval framework that integrates fine-grained object-level semantics with global scene-level spatial reasoning to enable context-aware, multimodal 3D asset retrieval. Unlike prior methods that only rely on object-centric cues (images or 3D shapes or text descriptions), MetaFind incorporates the spatial background by modeling the current scene layout as a structured graph. This layout-aware design allows the retriever to reason about placement constraints, positional dependencies, and contextual fit, enhancing spatial, semantic, and stylistic consistency. Moreover, MetaFind supports flexible multimodal queries, where the input can be any combination of text, image, point cloud, and layout context. This compositional design ensures robustness under missing modality conditions and adaptability to diverse use cases, including interactive scene editing, layout-conditioned asset generation, and large-scale virtual environment construction.

As shown in Figure\ref{fig:framework}, MetaFind builds upon ULIP2 \cite{xue2024ulip}, a tri-modal learning framework that aligns text, image, and point cloud into a shared embedding space. We adopt a dual-encoder architecture \cite{karpukhin2020dense}, where the query encoder flexibly encodes any user-provided modality combination, and the gallery encoder precomputes embeddings for all 3D assets to enable efficient retrieval. To supervise this alignment, we annotate 48K 3D assets from the Objaverse-LVIS subset \cite{deitke2023objaverse}, each rendered from 11 views and processed with GPT-4o to generate structured text descriptions. For layout-level reasoning, we introduce the Equivariant Spatial-Semantic Graph Neural Network (ESSGNN), an EGNN-based encoder designed to model rooms as graphs where nodes represent existing objects with 3D coordinates and text features and edges reflect spatial-semantic relationships. Unlike GNNs, ESSGNN maintains equivariance to rotation and translation by separating spatial and semantic channels, ensuring that scene embeddings remain stable across coordinate shifts and alignments—an essential property for robust layout modeling in unnormalized or dynamic environments. This encoder is trained on ProcTHOR \cite{deitke2022}, which contains over 10,000 generated houses. The ESSGNN outputs a layout context vector, which is fused with the query embedding to produce a layout-aware retrieval representation. We adopt a two-stage training: (1) pretraining on object-level data for cross-modal grounding and (2) fine-tuning on room-level scenes for layout-aware adaptation. This architecture ensures strong generalization, modularity, and robustness across complex retrieval conditions.

In summary, we contribute on: (1) we present \textbf{MetaFind}, a novel layout-aware multimodal 3D asset retrieval framework tailored for coherent scene generation, which jointly considers object-level features and scene-level spatial context; (2) we introduce a plug-and-play \textbf{ESSGNN} layout encoder that models the evolving scene as a structured graph, capturing spatial relationships, contextual dependencies, and semantic attributes to guide retrieval decisions, with built-in SE(3) equivariance to prevent degradation under arbitrary scene rotations or global shifts in coordinate systems; (3) we design MetaFind to support flexible and robust multimodal querying, allowing arbitrary combinations of multi-modalities as input, enabling strong performance under diverse and incomplete input conditions; and (4) we demonstrate through comprehensive experiments that MetaFind outperforms baselines in both standard retrieval and layout-aware scene construction, and that our proposed iterative retrieval pipeline enhances contextual consistency and realism compared to current methods.

% \vspace{-0.1in}
\section{Methodology}
% \vspace{-0.1in}
\input{2methdology}
% \vspace{-0.1in}
\input{3experiments}
% \vspace{-0.1in}
\section{Summary, Limitation, and Future Work}
% \vspace{-0.1in}
\label{sec:summary}
In this work, we present \textbf{MetaFind}, a scene-aware, multimodal 3D asset retrieval framework that unifies object-level semantics and scene-level spatial reasoning through a dual-tower design and a plug-and-play ESSGNN layout encoder. MetaFind demonstrates strong retrieval performance across both complete and partial modality settings, and significantly improves scene coherence and realism in iterative composition tasks. However, asset annotations rely on GPT-4o, which can introduce language bias, hallucinations, and occasional mislabeling (e.g., culturally skewed terms or incorrect attributes), potentially affecting training and evaluation. This work does not explicitly debias these annotations. Looking forward, we plan to extend MetaFind by incorporating real-world human-in-the-loop feedback for adaptive scene refinement, and scaling to open-world settings with dynamic object catalogs and evolving scene goals.

\clearpage

\section{Acknowledgments}
We gratefully acknowledge support from the NVIDIA Academic Grant (“Interactive Spatial Reasoning and 3D Scene Generation with RL-Enhanced VLMs”) and the provision of cloud computing resources, which enabled systematic training and evaluation of our MetaFind and other baselines. This paper is a core component of that project. The views expressed are those of the authors and do not necessarily reflect those of NVIDIA.
% \clearpage
\bibliographystyle{plain}
\bibliography{custom}

\clearpage
\input{appendix}

\end{document}

%% file: 2methdology.tex
\label{sec:methdology}
In this section, we introduce the MetaFind, formalize the retrieval task, and present our dual-tower architecture with modality-aware fusion and the ESSGNN layout encoder. We describe the training strategy and the iterative scene composition process for contextually coherent 3D asset retrieval.
% \vspace{-0.05in}
\subsection{Task Definition}
% \vspace{-0.05in}
We aim to accurately retrieve contextually coherent 3D assets from a large-scale repository, given a user query and optional existing scene layout information. Formally, our retrieval task can be defined as follows: given an input query \( Q = \{q_{text}, q_{img}, q_{pc}, q_{layout}\} \), which may include text \(q_{text}\), images \(q_{img}\), 3D point clouds \(q_{pc}\), and optionally layout context \(q_{layout}\), the system retrieves the asset \(A^*\) from a pre-encoded asset database \(\mathcal{A}\):
\begin{equation}
    A^* = \arg\max_{A \in \mathcal{A}} \text{sim}(f_{query}(Q), f_{gallery}(A)),
\end{equation}
where $f_{query}$ and $f_{gallery}$ represent the query and gallery encoders, and $\text{sim}(\cdot, \cdot)$ denotes the similarity function. The task is challenging due to the multimodal nature of user queries, partial modality absence, and the necessity for accurate layout awareness to ensure spatial coherence and realism.
% \vspace{-0.05in}
\subsection{Method Overview}
% \vspace{-0.05in}
To address the above challenge, we introduce MetaFind, as shown in \ref{fig:framework}, a dual-tower retrieval framework consisting of a query encoder and a gallery encoder, both leveraging the ULIP-2 embedding backbone. The gallery encoder precomputes embeddings for assets using three available modalities, which are then stored for efficient retrieval. On the query side, the encoder is designed to flexibly handle arbitrary combinations of modalities and, optionally, layout information—accommodating partial modality absence through a modality-aware fusion strategy. Specifically, each available modality is independently encoded using the ULIP-2 backbone, and these modality embeddings are subsequently integrated via a fusion layer, such as mean pooling, an MLP, or a Transformer-based module, generating a unified representation. Furthermore, the query encoder optionally integrates a layout encoder (ESSGNN) to capture spatial context from the existing scene layout. The layout is modeled as a structured graph with nodes representing placed objects (each described by spatial coordinates and semantic embeddings) and edges capturing spatial relationships. The layout encoder processes this graph to produce a context-aware layout vector, enhancing the spatial reasoning capability of the retrieval process. Its equivariant property ensure stable and generalizable scene embeddings under varying coordinate frames and unnormalized layouts common in open-world environments. 

Our training protocol involves two stages: First, we train the query and gallery encoders to learn fundamental multimodal embedding alignment without spatial context. Subsequently, we fine-tune the query encoder—particularly the fusion module and the layout encoder—using layout-aware room-level datasets. This fine-tuning stage employs adaptive freezing strategies, selectively freezing components like the gallery encoder to balance performance and computational efficiency.
% \vspace{-0.05in}
 \subsection{Data Preparation}
 % \vspace{-0.05in}
\begin{figure}[t]
    \centering
    \includegraphics[width=1\linewidth]{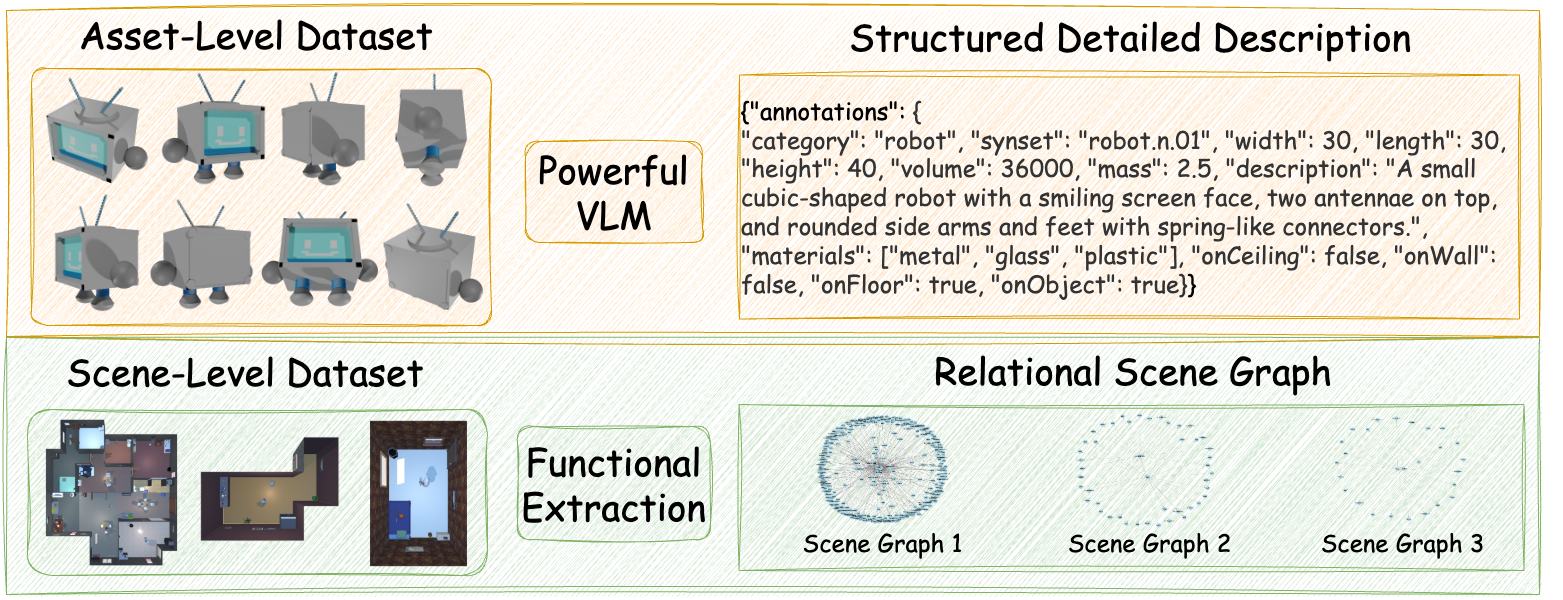}
    % \vspace{-0.25in}
    \caption{Data preparation pipeline. At the asset level (top), each 3D object from Objaverse-LVIS is rendered from multiple orthogonal views and passed through a VLM to generate structured, detailed annotations, capturing attributes such as category, dimensions, materials, and spatial placement constraints. At the scene-level (bottom), functional extraction is performed on generated rooms from the ProcTHOR, resulting in relational scene graphs encoding the spatial and semantic relationships between placed objects, enabling layout-aware retrieval capabilities in MetaFind.}
    \label{fig:data-sample}
    % \vspace{-0.2in}
\end{figure}
Our methodology requires prepared datasets at both object and scene levels to support multimodal and layout-aware retrieval tasks (as illustrated in Figure \ref{fig:data-sample}). For object-level representation learning, we utilize the Objaverse-LVIS dataset, which comprises approximately 48,000 distinct 3D assets. Each asset is rendered from 11 orthogonal viewpoints and annotated using GPT-4o. These annotations provide rich textual descriptions detailing attributes such as object category, size dimensions, materials, and placement constraints. For scene-level data, we leverage the ProcTHOR, which includes over 10,000 generated houses constructed from a curated collection of more than 3,000 unique assets. Each room configuration provides precise spatial coordinates and comprehensive semantic metadata for each asset, enabling the extraction of structured graphs representing object-level placements and spatial relationships. The bottom side of Figure \ref{fig:data-sample} illustrates the extraction process of such structured scene graphs. These graphs form the basis for training the layout-aware ESSGNN encoder, effectively capturing spatial coherence and relational context crucial for accurate asset retrieval. They include two types of edges: (i) physical-relation edges that capture spatial dependencies (e.g., “cup on table”); and (ii) semantic-relation edges that capture functional or contextual associations (e.g., “microscope–lab bench”), obtained by prompting an LLM on object pairs. This dual-edge design encodes both physical layout and high-level semantics, enhancing retrieval and layout reasoning.
% \vspace{-0.05in}
\subsection{Dual-Tower Architecture and Fusion Design}
% \vspace{-0.05in}
While prior works typically align 3D encoders to a fixed CLIP embedding space by freezing pretrained text and image encoders, our MetaFind framework adopts a more flexible dual-tower design. It enables context-aware, multi-modal queries by training a dedicated query encoder that fuses arbitrary modality subsets—including text, image, and scene-aware 3D inputs.

MetaFind employs a dual-tower architecture with separate encoders for the query and gallery. Each tower leverages ULIP-2 to independently encode available modalities (text, images, and point clouds). A modality-aware fusion module combines these modality embeddings via one of several strategies, such as mean pooling, MLP, masked MLP, gated fusion, or Transformer-based fusion. The gallery encoder is modality-complete and frozen after pretraining, while the query encoder remains flexible: It accepts any subset of modalities and can be augmented with a layout-aware vector. This vector is extracted using our proposed Equivariant Spatial-Semantic Graph Neural Network (ESSGNN) trained on scene graphs, enabling the model to incorporate spatial context for scene-aware retrieval.
% \vspace{-0.05in}
\subsection{ESSGNN: Scene-Aware Equivariant Graph Encoder}
% \vspace{-0.05in}
In this work, we propose the \textbf{Equivariant Spatial-Semantic Graph Neural Network (ESSGNN)} to encode 3D scene layouts in a way that is both spatially grounded and semantically expressive. ESSGNN is designed to maintain equivariance to SE(3) transformations during message passing while incorporating semantic relationships between objects through learned edge representations.

We initially experimented with standard Graph Attention Networks (GATs) to model inter-object dependencies based on spatial adjacency. However, we observed that GATs were highly sensitive to global translation and scaling variations across scenes, resulting in unstable layout embeddings and poor generalization. These issues are especially prominent in open-world or metaverse environments, where object positions are defined in large and often unnormalized coordinate systems, with no guarantee that scenes are aligned or centered.

Motivated by recent advances in drug design—where Equivariant Graph Neural Networks (EGNNs \cite{satorras2021n}) have been effectively applied to model 3D molecular structures invariant to spatial transformations—we design ESSGNN to address these limitations. Our model extends the EGNN formulation to incorporate semantic edge features in addition to geometric ones, allowing message passing to be informed not only by spatial proximity but also by functional or compositional relationships between objects. Given a scene graph \( G = (\mathcal{V}, \mathcal{E}) \), each node \( v_i \in \mathcal{V} \) represents an object with 3D position \( x_i \in \mathbb{R}^3 \) and a text-derived feature \( t_i \in \mathbb{R}^d \). The node feature is initialized as:
\[
    h_i^{(0)} = \text{Concat}(x_i, t_i).
\]

Edges in the graph include both spatial and semantic relationships. Spatial edges are extracted from physical layout constraints (e.g., adjacency, support), while semantic edges are generated by prompting a large language model (LLM) with object descriptions to produce natural language relation sentences. These sentences are then encoded into dense vectors using a frozen text encoder (e.g., CLIP or BERT), resulting in edge embeddings \( e_{ij} \) that carry functional and relational meaning.

The message-passing mechanism in ESSGNN follows a modified Equivariant Graph Convolutional Layer (EGCL) structure. For each layer \( l \), node features and positions are updated as:
\begin{align}
    h_i^{l+1} &= h_i^l + \sum_{j \in \mathcal{N}(i)} f_h(d_{ij}^l, h_i^l, h_j^l, e_{ij}; \theta_h), \\
    x_i^{l+1} &= x_i^l + \sum_{j \in \mathcal{N}(i)} (x_i^l - x_j^l) \cdot f_x(d_{ij}^l, h_i^{l+1}, h_j^{l+1}, e_{ij}; \theta_x),
\end{align}
where \( d_{ij}^l = \|x_i^l - x_j^l\|_2 \) is the Euclidean distance between nodes, and \( f_h: \mathbb{R}^{(2d + 1 + e)} \to \mathbb{R}^d \), \( f_x: \mathbb{R}^{(2d + 1 + e)} \to \mathbb{R}^{3} \) are two learnable functions parameterized by \( \theta_h \) and \( \theta_x \), respectively, which we approximate using multilayer perceptrons (MLPs). Here, \( e \) denotes the dimension of the semantic edge embedding \( e_{ij} \). After \( L \) layers, the node features are aggregated into a global layout embedding:
\[
    e_{\text{layout}} = \text{Pooling}(\{h_i^{(L)}\}).
\]

This embedding is integrated into the query encoder of our dual-tower retrieval framework to provide scene-aware conditioning. ESSGNN generalizes the original EGNN by introducing \textit{semantic-aware edge modulation}, enabling it to operate on multi-relational graphs with heterogeneous object types and to better handle complex spatial-functional layouts found in real-world and virtual 3D scenes.

Our model retains full SE(3)-equivariance concerning input transformations. Specifically, for any rotation operator \( R \in SO(3) \) and translation vector \( T \in \mathbb{R}^3 \), the following condition holds:
\begin{align}
    (R x^{l+1} + T,\, h^{l+1}) = \text{ESSGNN}(R x^l + T,\, h^l,\, E). \label{eq:EGCL_equivariance}
\end{align}
We provide a formal proof of this equivariance property in Appendix~\ref{sec:appendix_equiv_proof}, extending the original EGCL proof to include semantic edge features.

% \vspace{-0.05in}
\subsection{Training Strategy}
% \vspace{-0.05in}
We adopt a two-stage training strategy that aligns with the dual-tower architecture and the flexible, multimodal nature of the retrieval task. In the first stage, we focus on learning robust cross-modal representations that can handle arbitrary combinations of query modalities. In the second stage, we incorporate scene layout information through an ESSGNN encoder, enabling the system to perform context-aware retrieval grounded in spatial reasoning.
% \vspace{-0.1in}
\subsubsection*{Stage 1: Cross-Modal Alignment Pretraining.}
% \vspace{-0.05in}
In the first stage, both query and gallery encoders are trained on large-scale object-level data from Objaverse-LVIS, where each asset has full modality inputs (text, images, and point clouds). We introduce stochastic modality masking to simulate partial-modality queries: each modality in the query has a 30\% probability of being independently masked. Rather than zero-padding, we apply masked embeddings to ensure flexibility and prevent model degradation. The goal is to align all available modality combinations into a shared embedding space. The gallery encoder is trained to be modality-complete, and both towers share the contrastive retrieval objective:
\begin{equation}
    \mathcal{L}_{\text{pre}} = -\log \frac{\exp(\text{sim}(f_{\text{query}}(Q), f_{\text{gallery}}(A)) / \tau)}{\sum_{A' \in \mathcal{B}} \exp(\text{sim}(f_{\text{query}}(Q), f_{\text{gallery}}(A')) / \tau)},
\end{equation}
where \( \tau \) is a temperature hyperparameter and \( \mathcal{B} \) denotes the gallery batch. 
% \vspace{-0.1in}
\subsubsection*{Stage 2: Layout-Aware Fine-Tuning}
% \vspace{-0.05in}
In the second training stage, we enhance the query encoder with spatial context derived from the current scene layout. Given available modality embeddings for text \(e_{\text{text}}\), image \(e_{\text{img}}\), and point cloud \(e_{\text{pc}}\), along with the optional layout embedding \(e_{\text{layout}}\) produced by the ESSGNN module, the final fused query representation is computed as:
\begin{equation}
    e_{\text{query}} = \text{Fusion}(e_{\text{text}}, e_{\text{img}}, e_{\text{pc}}) + \lambda \cdot e_{\text{layout}},
\end{equation}
where \(\lambda\) is a learnable scalar controlling the contribution of layout information. This residual design allows layout reasoning to enhance retrieval without disrupting the original embedding space.

To ensure robustness in real-world settings where scene layouts may not always be available, we introduce \textit{stochastic scene dropout} during training: the layout vector \(e_{\text{layout}}\) is omitted in 30\% of batches, forcing the model to generalize to layout-free inputs. Only the query-side fusion layer and the ESSGNN module are updated during this stage; the gallery encoder is frozen to reduce training costs and preserve asset embedding consistency.

We adopt a \textbf{bidirectional contrastive learning} objective to symmetrically align query and gallery embeddings. Let \(e_{\text{query}}\) and \(e_{\text{gallery}}\) denote the fused query and gallery embeddings, respectively. The layout-aware retrieval loss is defined as:
{\small
\begin{equation}
    \mathcal{L}_{\text{layout}}^{\text{q2g}} = -\log \frac{\exp(\text{sim}(e_{\text{query}}, e_{\text{gallery}}) / \tau)}{\sum_{e'_{\text{gallery}} \in \mathcal{B}} \exp(\text{sim}(e_{\text{query}}, e'_{\text{gallery}}) / \tau)}, \quad
    \mathcal{L}_{\text{layout}}^{\text{g2q}} = -\log \frac{\exp(\text{sim}(e_{\text{gallery}}, e_{\text{query}}) / \tau)}{\sum_{e'_{\text{query}} \in \mathcal{B}} \exp(\text{sim}(e_{\text{gallery}}, e'_{\text{query}}) / \tau)}
\end{equation}
}

where \(\tau\) is a temperature hyperparameter, and \(\mathcal{B}\) denotes the batch of negatives. The final loss is the average of the two directions:
\begin{equation}
    \mathcal{L}_{\text{layout}} = \frac{1}{2} \left( \mathcal{L}_{\text{layout}}^{\text{q2g}} + \mathcal{L}_{\text{layout}}^{\text{g2q}} \right).
\end{equation}

This training strategy encourages accurate retrieval of relevant assets (query-to-gallery) and consistent representation of assets retrievable by matching scene context (gallery-to-query). The model improves generalization and robustness by aligning both directions, especially in iterative scene construction where queries and context evolve.

% \vspace{-0.05in}
\subsection{Inference and Iterative Composition}
% \vspace{-0.05in}
At inference time, all gallery asset embeddings are precomputed and cached for efficient retrieval. Given an input query—which may consist of any combination of text, image, point cloud, and optional scene layout—the query encoder generates a layout-aware embedding used to identify the most contextually suitable asset from the gallery. 

To construct complete scenes, we deploy an iterative composition strategy shown in Algorithm~\ref{alg:iterative}. Instead of retrieving all required objects independently in a single step, we retrieve and place one object at a time. After each placement, the scene graph is updated to reflect the new layout, and the ESSGNN module recomputes the layout embedding, allowing subsequent retrievals to account for the evolving spatial context. While this step-by-step process introduces additional computational latency compared to one-shot parallel retrieval, it significantly improves spatial coherence and contextual alignment across placed objects, resulting in more realistic and visually harmonious scenes.

\textbf{Efficiency considerations.} The iterative pipeline incurs extra latency and compute versus one-shot retrieval—especially for multi-object scenes—but this trade-off is use-case dependent and tunable. When global coherence and stylistic consistency matter most, a fully sequential schedule yields the best quality. When efficiency is prioritized, we use parallel retrieval or region-based decomposition: partition a room into semantic/spatial regions (e.g., seating, storage), retrieve sequentially within each region to preserve local coherence, and process regions in parallel to improve throughput. This design flexibility makes the method practical across scenarios, and we have clarified it in the revision.

\begin{algorithm}[t]
\caption{Iterative Layout-Aware Scene Composition}
\label{alg:iterative}
\begin{algorithmic}[1]
\REQUIRE Precomputed gallery embeddings $\mathcal{E}_{\text{gallery}}$, initial scene graph $G_0$, asset query list $\{Q_1, Q_2, ..., Q_N\}$

\STATE Initialize scene graph $G \leftarrow G_0$

\FOR{$i = 1$ to $N$}
    \STATE Extract current layout embedding: $e_{\text{layout}} \leftarrow \text{EGNN}(G)$
    \STATE Encode available modalities of query $Q_i$: $e_{\text{text}}, e_{\text{img}}, e_{\text{pc}}$
    \STATE Fuse into layout-aware query: $e_{\text{query}} \leftarrow \text{Fusion}(e_{\text{text}}, e_{\text{img}}, e_{\text{pc}}) + \lambda \cdot e_{\text{layout}}$
    \STATE Retrieve best-matching asset: $A^*_i \leftarrow \arg\max_{A \in \mathcal{E}_{\text{gallery}}} \text{sim}(e_{\text{query}}, e_{\text{gallery}}(A))$
    \STATE Place $A^*_i$ into the scene, update scene graph: $G \leftarrow G \cup \{A^*_i\}$
\ENDFOR

\STATE \textbf{return} Final composed scene $G$
\end{algorithmic}
\end{algorithm}

%% file: 3experiments.tex
\section{Experiments}
\label{sec:experiments}
% \vspace{-0.1in}
We conduct comprehensive experiments to evaluate MetaFind across multiple dimensions, including object-level retrieval, scene-level layout-aware retrieval, and robustness under varying design choices. We begin by introducing our experimental setup, datasets, and baseline adaptations. We then present quantitative results on the Objaverse-LVIS dataset to assess retrieval performance under different modality combinations. Next, we evaluate scene-level quality on the ProcTHOR dataset, highlighting the benefits of layout-aware retrieval using our ESSGNN context encoder. We further perform extensive ablation studies to analyze the contribution of core architectural components and training strategies. Finally, we assess generalization across scene complexities and provide qualitative visualizations to showcase the real-world effectiveness of MetaFind.
% \vspace{-0.05in}
\subsection{Experimental Setup}
% \vspace{-0.05in}
\paragraph{Datasets} We evaluate MetaFind across both object-level and scene-level retrieval settings. The object-level experiments are conducted on the annotated Objaverse-LVIS dataset containing 48K unique 3D assets. For scene-level layout-aware retrieval, we use the ProcTHOR-10K dataset containing over 10,000 procedurally generated house layouts constructed from over 3,000 curated 3D assets. In both datasets, we allocate 80\% of the data for training and reserve the remaining 20\% for testing. While our experiments currently use single-room indoor scenes, the framework is designed to generalize to open-world settings; the $\mathrm{SE}(3)$-equivariant design specifically targets robustness to large-scale and dynamic environments.
% \vspace{-0.05in}
\paragraph{Baselines} 
\textbf{ULIP}~\cite{xue2024ulip} is a tri-modal single-tower model that aligns text, image, and point cloud modalities into a unified embedding space through joint representation learning. \textbf{OpenShape}~\cite{liu2023openshape} adopts a dual-tower contrastive retrieval design, supporting text-to-3D and image-to-3D retrieval via large-scale vision-language pretraining. \textbf{SCA3D}~\cite{ren2025sca3d} focuses on point cloud-text retrieval and improves robustness using self-augmented contrastive learning, though it lacks multi-modal query fusion capabilities. 
% \textbf{COM3D}~\cite{wu2024com3d} is a compositional retrieval model that incorporates spatial priors and semantic cues for improved category-level 3D asset retrieval from textual input. 
\textbf{Uni3DL}~\cite{li2023uni3dl} and \textbf{Uni3D}~\cite{zhou2023uni3d} present unified architectures for 3D-language-image understanding, supporting multiple modalities inputs. Finally, \textbf{OmniBind}~\cite{wang2024omnibindlargescaleomnimultimodal} offers a scalable omni-modality representation space that supports combinations of text, image, audio, and point cloud inputs, though it is not optimized for layout-aware or scene-conditioned retrieval.

Since most existing retrieval models (e.g., ULIP, OpenShape) are not designed to handle arbitrary combinations of input modalities, we limit our baselines to pre-trained single-tower encoders that support at least one of the three modalities: text, image, and point cloud. To create a fair comparison within a dual-tower retrieval paradigm, we extend each baseline by adding a simple \textit{mean pooling layer} to aggregate available modalities, and use these fused embeddings to retrieve from a pre-encoded gallery. For completeness, we also include our own dual-tower model with a mean fusion layer but without layout context as a direct ablation baseline. The temperature is 0.5 for all experiments.

% \vspace{-0.05in}
\paragraph{Metrics} We benchmark MetaFind and all variants using standard retrieval metrics, including top-$k$ retrieval accuracy (R@1, R@5). To assess scene-level performance, we further evaluate the compositional quality of generated scenes along two axes: structural coherence and stylistic consistency. These aspects are quantitatively scored using a GPT-4o-based aesthetic and alignment evaluator, and qualitatively validated through human preference studies conducted on a subset of generated scenes. This dual evaluation setup provides a comprehensive assessment of both retrieval accuracy and real-world usability in downstream scene construction.

% \vspace{-0.05in}
\subsection{Retrieval Performance on Objaverse-LVIS}
% \vspace{-0.05in}
We first evaluate the object-level retrieval performance on the annotated Objaverse-LVIS dataset, which comprises 48K high-quality 3D assets with structured textual descriptions and multi-view image renders. This evaluation focuses on the core capability of MetaFind to support flexible, modality-compositional retrieval, especially under partial modality conditions. All methods are evaluated under seven query conditions: text-only, image-only, point cloud-only, text+image, text+point cloud, image+point cloud, and full (text+image+point cloud). As shown in Table~\ref{tab:objaverse-results}, MetaFind without ESSGNN outperforms all baseline models across different settings. Notably, since other models do not adopt a dual-tower design, their "PC only" performance reflects retrieval using identical embeddings for both query and gallery, leading to inflated accuracy. In contrast, our dual-tower framework introduces more cross-modality retrieval, which results in lower accuracy under the "PC only". Nevertheless, MetaFind demonstrates stronger performance under partial modality conditions, highlighting its capability in multimodal fusion. After integrating the ESSGNN, while the overall scene quality is improved, we observe a drop in accuracy due to the added encoded information. This reflects a temporary and explainable trade-off between object-level precision and scene-level coherence. Stage-1 pretraining on Objaverse-LVIS uses isolated assets (no layout) and no ESSGNN; Stage-2 fine-tuning introduces ESSGNN on ProcTHOR (layout-rich, different asset distribution). Although the retrieval objective is unchanged, the fusion layer becomes partially adapted to layout-conditioned features, creating a feature-attribution mismatch when evaluating on Objaverse-LVIS (which lacks layout and disables ESSGNN). A practical mitigation is to maintain two fusion heads: a layout-free head (Stage-1) and a scene-aware head (Stage-2), selected at inference by context availability. Using the Stage-1 head reproduces the “w/o ESSGNN” numbers (omitted for brevity). In our reported results, we instead explore a single shared head by freezing both encoders in Stage-2, updating only ESSGNN and the fusion, and applying stochastic scene dropout (30\%) to expose the model to layout-free inputs; some accuracy loss remains due to residual attribution drift.

\begin{table}[h]
\centering
% \vspace{-0.2in}
\caption{Retrieval accuracy (R@1 / R@5) on Objaverse-LVIS under different query modality combinations. MetaFind consistently outperforms all baselines across both complete and incomplete query settings. `--` indicates that the method does not support the corresponding modality combination.}
\label{tab:objaverse-results}
\resizebox{\textwidth}{!}{%
\begin{tabular}{lccccccc}
\toprule
\textbf{Method} & \textbf{Text Only} & \textbf{Image Only} & \textbf{PC Only} & \textbf{T + I} & \textbf{T + PC} & \textbf{I + PC} & \textbf{T + I + PC} \\
\midrule
ULIP~\cite{xue2024ulip} & 0.1 / 0.9 & 0.1 / 1.3 & 97.9 / 99.4 & 0 / 0.3  & 33.9 / 58 & 22.6 / 41.6 & 6.4 / 15.9 \\
OpenShape~\cite{liu2023openshape} &  0.6 / 1.7 & 0.3 / 1.1 & 98.4 / 99.7 & 0 / 0.5 & 35.1 / 61.4 & 25.0 / 44.3 & 7.0 / 17.2 \\
SCA3D~\cite{ren2025sca3d} & 6.9 / 10.4 & -- & 98.1 / 99.3 & -- & 39.7 / 65.2 & -- & -- \\
% COM3D~\cite{wu2024com3d} & X / X & X / X & X / X & X / X & X / X & X / X & X / X \\
Uni3DL~\cite{li2023uni3dl} & 4.5 / 9.2 & -- & \underline{98.5} / \textbf{99.8} & -- & 37.4 / 63.9 & -- & -- \\
Uni3D~\cite{zhou2023uni3d} & 1.7 / 3.9 & 1.2 / 2.5 & 98.3 / 99.4 & 0.5 / 1.1 & 36.3 / 63.6 & 26.1 / 44.8 & 8.2 / 19.1 \\
OmniBind (Base) & 1.2 / 2.8 & 0.6 / 1.4 & 98.3 / 99.6 & 0 / 0.4 & 34.0 / 55.9 & 21.5 / 38.7 & 5.5 / 13.8 \\
OmniBind (Large) & 2.7 / 4.0 & 0.9 / 1.8 & 98.2 / 99.3 & 0.1 / 0.4 & 35.2 / 56.7 & 23.4 / 40.9 & 6.0 / 16.7 \\
OmniBind (Full)\cite{wang2024omnibindlargescaleomnimultimodal} & 5.3 / 11.7 & 2.3 / 3.5 & \textbf{99.0} / \underline{99.7} & 0.5 / 1.2 & 37.5 / 60.8 & 27.5 / 46.4 & 11.9 / 23.4 \\
MetaFind w/o ESSGNN & \textbf{13.8 / 23.1} & \textbf{11.7 / 19.2} & 75.1 / 78.0 & \textbf{17.2 / 21.8} & \textbf{44.5 / 71.3} & \textbf{45.8 / 73.1} & \textbf{51.7 / 76.5} \\
MetaFind w/ ESSGNN & \underline{11.3 / 21.5} &  \underline{10.5 / 15.9}  & 63.2 / 66.5 & \underline{15.9 / 20.3} & \underline{41.2 / 68.8} & \underline{42.0 / 70.4} & \underline{48.2 / 74.9}\\ 
\bottomrule
\end{tabular}%
}
% \vspace{-0.15in}
\end{table}
% \vspace{-0.05in}
\subsection{Scene-Level Retrieval with Layout Context}
% \vspace{-0.05in}
To evaluate the benefit of layout-aware retrieval in realistic scenes, we assess MetaFind on a scene generation pipeline of I-Design \cite{ccelen2024design}. It can generate a 3D scene with a given room description by designing, retrieving, and arranging.  In the original paper, they use OpenShape\cite{liu2023openshape} to retrieve the objects. Here, we compare the performance of MetaFind with and without the ESSGNN layout encoder. No retrieval accuracy, we assess the overall quality of composed scenes using both automated and human evaluations across four key dimensions: (1) Overall Aesthetic and Atmosphere: Measures the visual appeal and mood of the composed scene; (2) Color Scheme and Material Choices: Evaluates consistency in textures, colors, and materials between newly retrieved assets and the existing scene; (3) Scene Coherence: Assesses how well the inserted assets align with the scene's spatial and semantic context; and (4) Realism and 3D Geometric Consistency: Checks for physically plausible placements, avoiding collisions or unnatural geometry. Each dimension is rated on a scale from 1 (poor) to 5 (excellent), independently by GPT-4o and five expert human annotators on a set of 200 randomly sampled scenes. For GPT-4o, we provide scene layouts and rendered views, with prompts aligned to the respective evaluation criteria. Final scores are averaged across annotators and samples.

\begin{figure*}[t]
  \centering
  \begin{subfigure}[b]{0.23\textwidth}
    \centering
    \includegraphics[width=3.5cm, height=3.5cm]{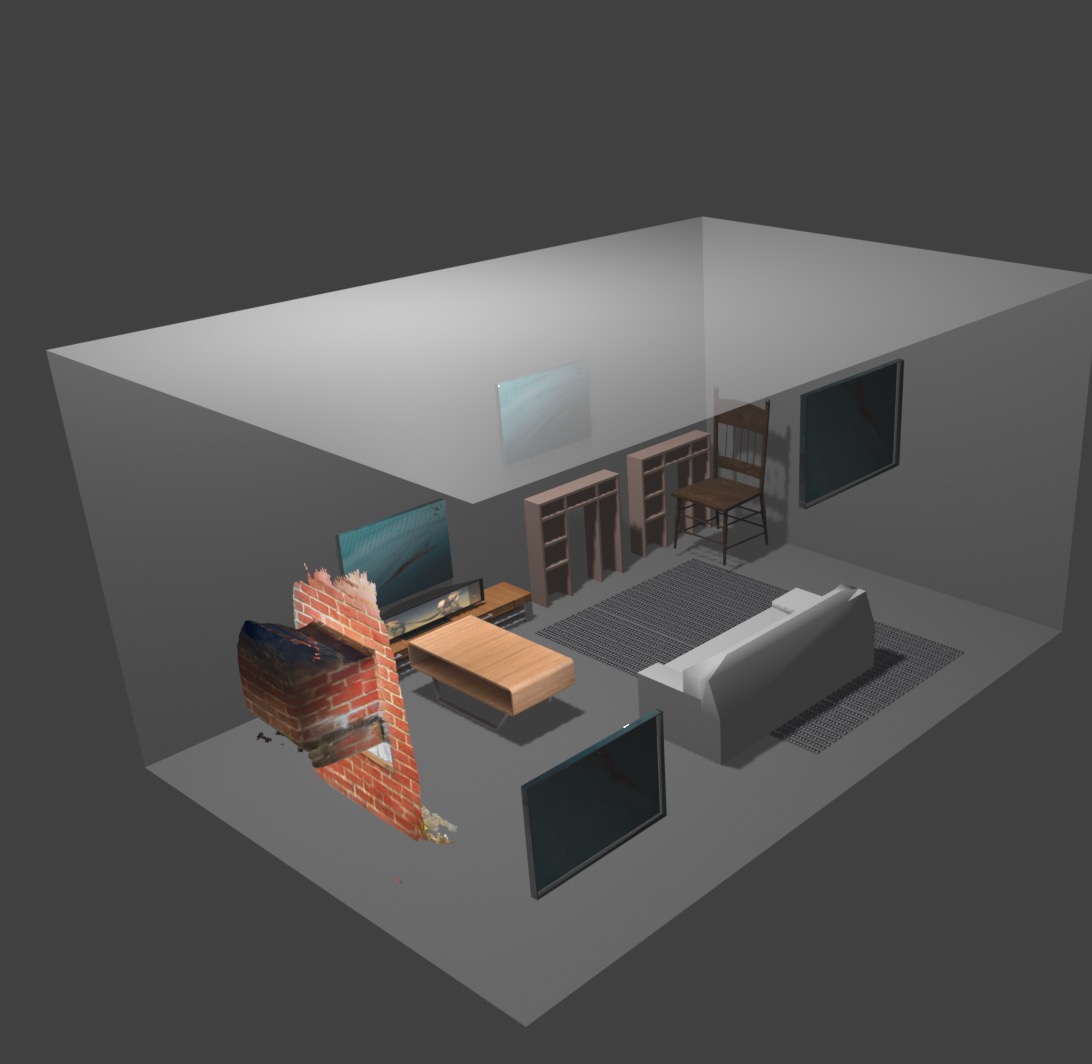}
    \caption*{Room 1\\Without ESSGNN}
  \end{subfigure}
  \begin{subfigure}[b]{0.23\textwidth}
    \centering
    \includegraphics[width=3.5cm, height=3.5cm]{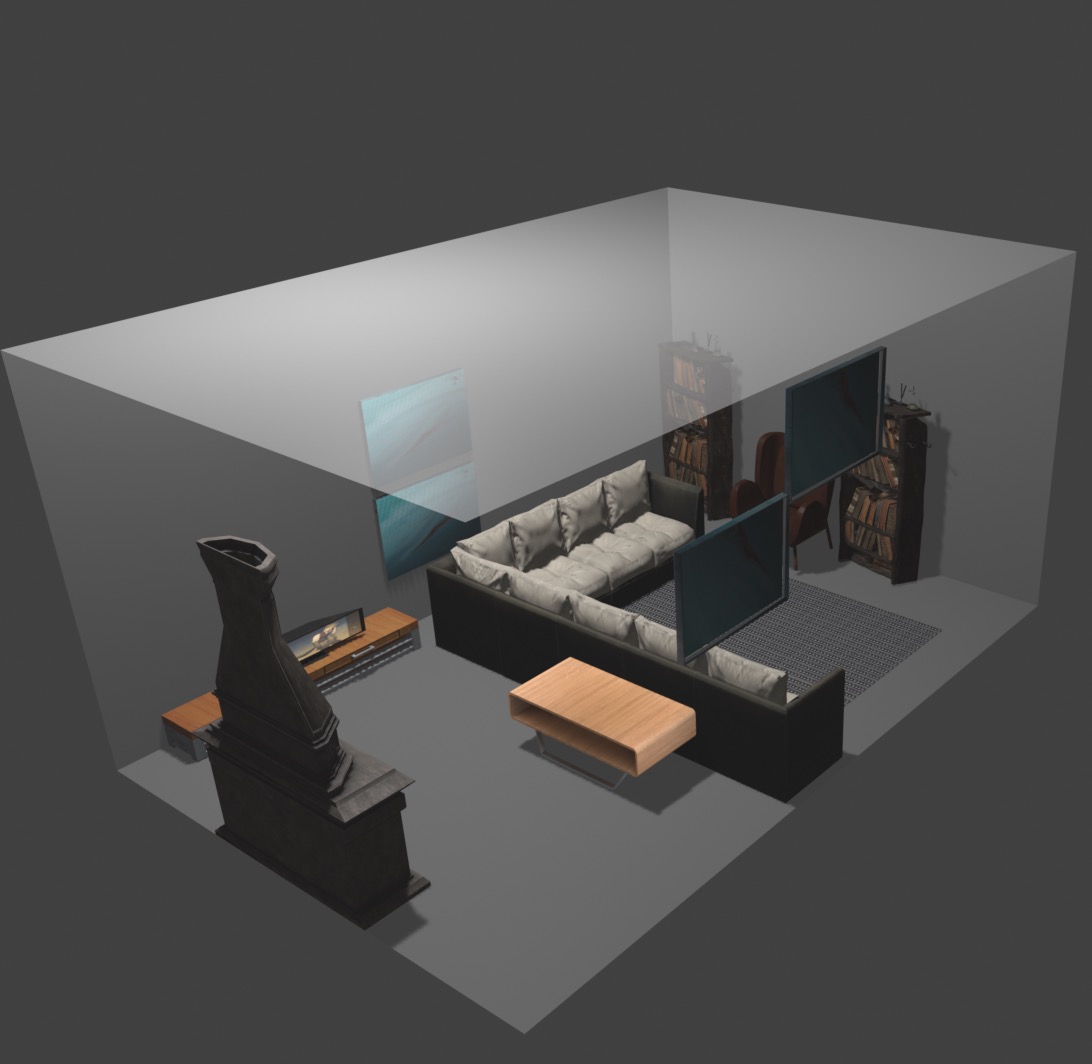}
    \caption*{Room 1\\With ESSGNN}
  \end{subfigure}
  \begin{subfigure}[b]{0.23\textwidth}
    \centering
    \includegraphics[width=3.5cm, height=3.5cm]{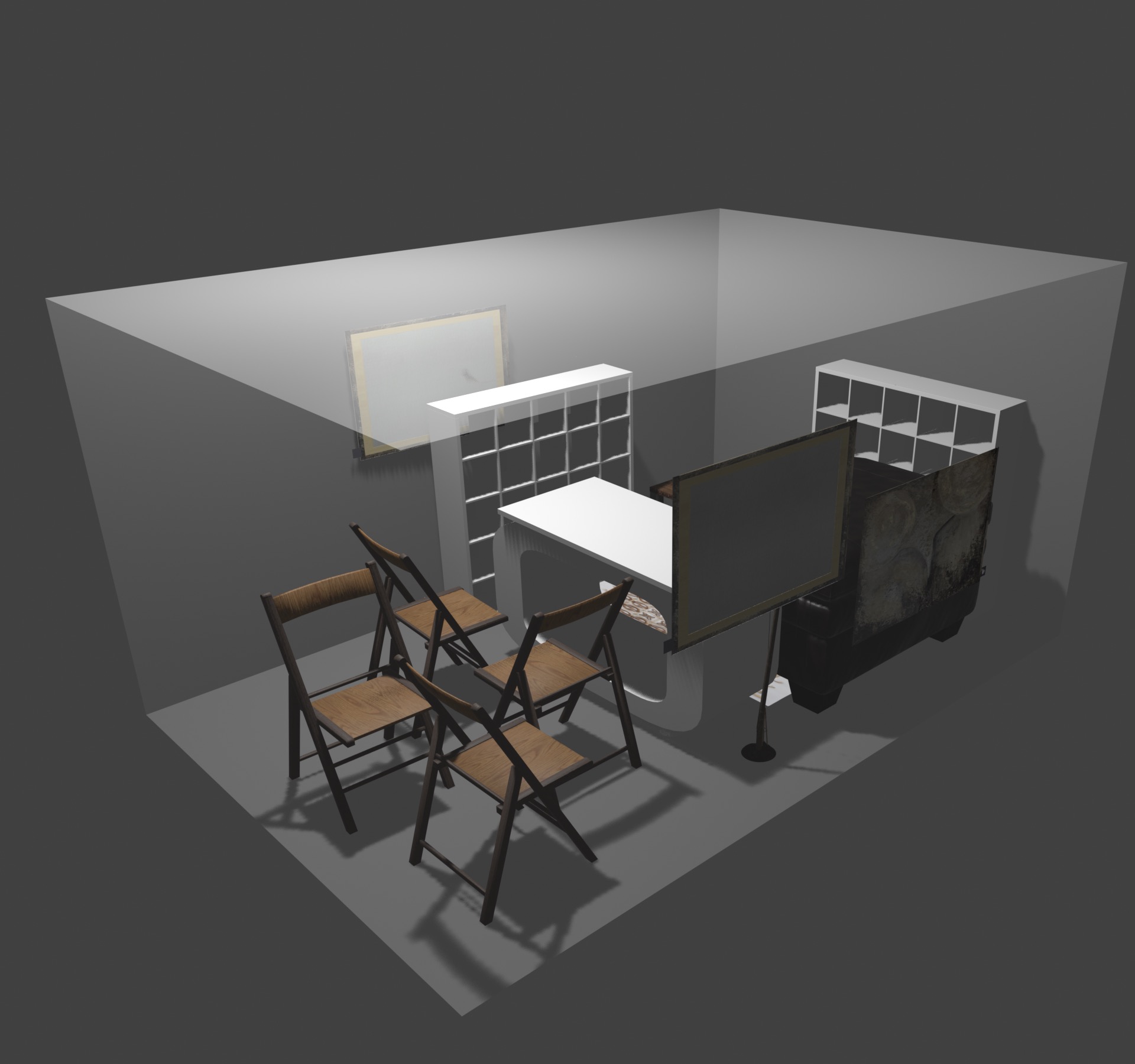}
    \caption*{Room 2\\Without ESSGNN}
  \end{subfigure}
  \begin{subfigure}[b]{0.23\textwidth}
    \centering
    \includegraphics[width=3.5cm, height=3.5cm]{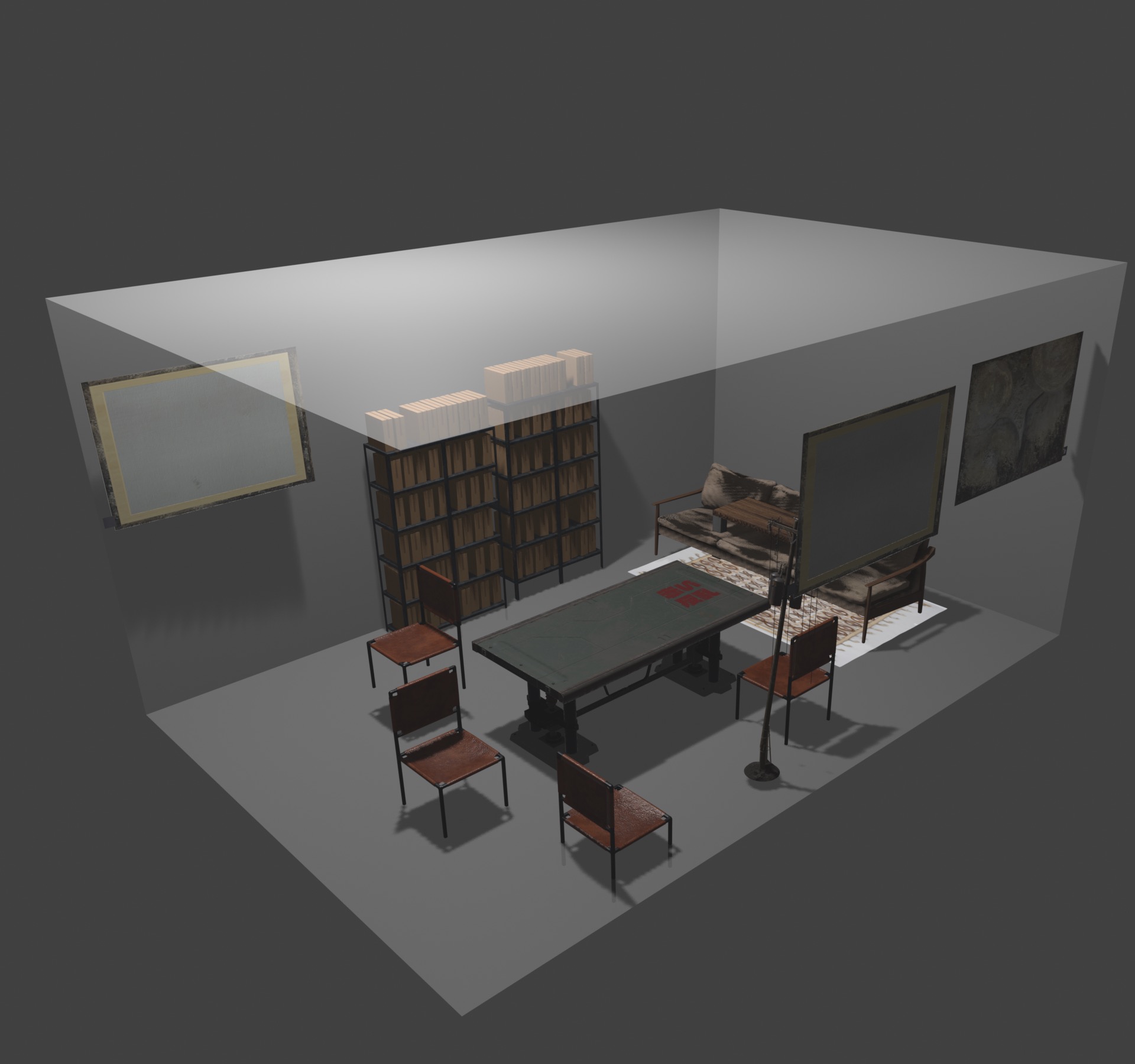}
    \caption*{Room 2\\With ESSGNN}
  \end{subfigure}

  \caption{Visual comparison of scene generation with and without the ESSGNN encoder across two room descriptions. Room 1 — "A classical-style lounge for group leisure and conversation"; Room 2 — “An aged archive room for research and consultation”}
  \label{fig:essgnn_row4}
  % \vspace{-0.15in}
\end{figure*}

\begin{table}[h]
\centering
% \vspace{-0.15in}
\caption{Scene-level quality comparison across four evaluation dimensions. MetaFind (with GSSNN) achieves the highest scores across both GPT-4o and human evaluations, demonstrating superior spatial coherence and aesthetic quality in composed scenes.}
\label{tab:layout-results}
\resizebox{\textwidth}{!}{%
\begin{tabular}{lcccccccc}
\toprule
\multirow{2}{*}{\textbf{Method}} & \multicolumn{2}{c}{\textbf{Aesthetic}} & \multicolumn{2}{c}{\textbf{Color \& Material}} & \multicolumn{2}{c}{\textbf{Scene Coherence}} & \multicolumn{2}{c}{\textbf{Realism \& Geometry}} \\
\cmidrule(lr){2-3} \cmidrule(lr){4-5} \cmidrule(lr){6-7} \cmidrule(lr){8-9}
 & GPT-4o & Human & GPT-4o & Human & GPT-4o & Human & GPT-4o & Human \\
\midrule
ULIP~\cite{xue2024ulip} & 2.91 & 3.02 & 2.84 & 2.97 & 2.76 & 2.89 & 2.70 & 2.81 \\
OpenShape~\cite{liu2023openshape} & 3.14 & 3.28 & 3.08 & 3.19 & 3.01 & 3.11 & 2.95 & 3.06 \\
MetaFind w/o ESSGNN & \underline{3.42} & \underline{3.55} & \underline{3.31} & \underline{3.41} & \underline{3.26} & \underline{3.33} & \underline{3.22} & \underline{3.30} \\
MetaFind w/ ESSGNN & \textbf{4.13} & \textbf{4.25} & \textbf{4.04} & \textbf{4.17} & \textbf{4.10} & \textbf{4.21} & \textbf{4.06} & \textbf{4.18} \\
\bottomrule
\end{tabular}
}
% \vspace{-0.1in}
\end{table}
Figures~\ref{fig:essgnn_row4}, \ref{fig:essgnn_comparison} show qualitative comparisons of scene generation with and without the ESSGNN encoder. The first example is a classical-style lounge, which, without ESSGNN, suffers from inconsistent object styles and poor layout organization. With ESSGNN, the scene is more coherent, with well-aligned furniture and logical arrangement for group interaction. The second example is an aged archive room. Without ESSGNN, the objects appear mismatched, while the ESSGNN-generated version offers a more functional and visually consistent space, with well-placed furniture suitable for a reading environment. These results demonstrate that ESSGNN improves both stylistic consistency and layout functionality. This qualitative improvement is also reflected in the quantitative results shown in Table~\ref{tab:layout-results}, where MetaFind with ESSGNN achieves the highest scores across all evaluation metrics. In particular, the gains in scene coherence and realism highlight the encoder’s ability to model spatial relationships and stylistic alignment effectively. Together, these findings confirm the effectiveness of ESSGNN in generating high-quality, semantically grounded 3D scenes.

% \vspace{-0.05in}
\subsection{Ablation Studies}
% \vspace{-0.05in}
\begin{table}[h]
\centering
% \vspace{-0.1in}
\caption{Ablation study (Text Only). We report top-1 retrieval accuracy (R@1) on the Object-level task, GPT-4o-based aesthetic score, and scene-level coherence score on the Scene-Level task.}
\label{tab:ablation}
\resizebox{\textwidth}{!}{
\begin{tabular}{lccc}
\toprule
\textbf{Variant} & \textbf{R@1 (\%)} & \textbf{Aesthetic (GPT-4o)} & \textbf{Scene Coherence (GPT-4o)} \\
\midrule
\textbf{MetaFind (Full, bidirectional) w/ iterative retrieval \& ESSGNN} & 11.4 & \textbf{4.1} & \textbf{4.2} \\
\midrule
w/o iterative retrieval & 11.3 & \underline{4.0} & \underline{4.1} \\
w/o Layout Context & \textbf{13.5} & 3.4 & 3.3 \\
w/ Layout Context (GAT) & 11.0 & 3.4 & 3.7 \\
\midrule
Fusion = Mean & 9.4 & 3.2 & 3.5 \\
Fusion = MLPs & 9.9 & 3.3 & 3.5 \\
\midrule
Modality Dropout = 10\% & 7.3 & 3.4 & 3.5 \\
Modality Dropout = 50\% & \underline{13.2} & 3.1 & 3.2 \\
\midrule
Train fuser only & 8.7 & 3.3 & 3.2 \\
\midrule
Padding missing modalities with 0& 10.5 & 3.1 & 3.1 \\
\bottomrule
\end{tabular}
}
% \vspace{-0.1in}
\end{table}

We conduct ablation studies to evaluate the effectiveness of key architectural components and training strategies in MetaFind, focusing on six dimensions: layout encoding, modality fusion strategies, modality dropout robustness, fusion granularity, gallery encoder flexibility, and missing modality handling. First, removing the ESSGNN layout encoder results in drops in scene realism, underscoring the critical role of spatial context. Regarding fusion strategies, while simple mean pooling offers computational efficiency, MLP and the final selected Transformer outperform others under partial modality conditions by dynamically reweighting available inputs. We also examine modality dropout rates during training, finding that a 30\% rate strikes the best balance between robustness and accuracy. Lower rates lead to overfitting on full-modality inputs, whereas higher rates introduce instability. Additionally, we compare fusion granularity strategies, revealing that while training only the fusion module in the query encoder improves efficiency, full encoder fine-tuning yields better performance by allowing earlier layers to adapt to modality-aware supervision. Finally, in handling missing modalities, modality masking outperforms zero-padding by preventing zero embedding interference and promoting robustness through sparsity-aware fusion. Results across these ablations, summarized in Table~\ref{tab:ablation}, demonstrate the modularity and resilience of MetaFind under diverse design choices.

%% file: appendix.tex
\appendix
\part*{Appendix}
{
\setlength{\parskip}{-0em}
\startcontents[sections]
\printcontents[sections]{ }{1}{}
}

\label{sec:appendix}

\section{Borader Impacts}
\label{sec::broader}
MetaFind facilitates accessible and coherent 3D scene generation, which can benefit fields like virtual reality, education, and game design. By supporting flexible multimodal queries, it lowers the barrier for non-experts to build rich virtual environments. However, risks include potential misuse in generating misleading content, propagation of bias from training data, and intellectual property concerns tied to retrieved assets. We recommend responsible dataset curation and human oversight to ensure ethical deployment.
\input{4backgound}

\section{Equivariance Proof of ESSGNN - Extension to Semantic Embedding} \label{sec:appendix_equiv_proof}

In this section, we prove that our ESSGNN maintains SE(3) equivariance in 3D space. While the original EGNN \cite{satorras2021n} formulation allows the inclusion of edge features in the message function, these are typically discrete, task-specific features such as bond types or edge labels. In contrast, our ESSGNN introduces edge embeddings \( e_{ij} \) derived from LLM-generated natural language relation descriptions, which are subsequently encoded via a frozen text encoder. Importantly, these semantic edge embeddings are \textit{invariant to the input node positions} \( \rm x \), as they are computed solely from object-level text descriptions and do not depend on spatial coordinates. Therefore, although the semantics encoded in \( e_{ij} \) are richer and more expressive, the mathematical property required for equivariance—the independence of \( e_{ij} \) from \( \rm x \)—remains satisfied. As a result, the message and update equations remain SE(3)-equivariant under our semantic extension, and the original proof structure holds. We now restate and extend the proof below.

Specifically, we show that for any translation vector \( g \in \mathbb{R}^3 \) and any orthogonal transformation \( Q \in \mathbb{R}^{3 \times 3} \), the model satisfies:

\begin{equation}
    Q\rm x^{l+1} + g,\; \rm h^{l+1} = \mathrm{ESSGNN}(Q\rm x^l + g,\; \rm h^l,\; E)
\end{equation}

where \( \rm x^l \) and \( \rm h^l \) are the positions and features of all nodes at layer \( l \), and \( E \) contains edge features including learned semantic embeddings \( e_{ij} \). We begin by assuming that \( \rm h^0 \) is invariant to SE(3) transformations on \( \rm x \), and that semantic edge embeddings \( e_{ij} \) are derived solely from object-level textual descriptions and thus independent of spatial coordinates. Under these assumptions, the edge message computation remains SE(3) invariant. Let us denote the pairwise edge message as:

\begin{equation}
    \rm m_{ij} = \phi_e\left(\rm h_i^l, \rm h_j^l, \| \rm x_i^l - \rm x_j^l \|^2, e_{ij} \right)
\end{equation}
    
Now consider a translation and rotation of all node positions: \( \rm x_i^l \mapsto Q\rm x_i^l + g \). The Euclidean distance term becomes:

\begin{equation}
    \| Q\rm x_i^l + g - (Q\rm x_j^l + g) \|^2 = \| Q(\rm x_i^l - \rm x_j^l) \|^2 = \| \rm x_i^l - \rm x_j^l \|^2
\end{equation}

Hence the edge message is preserved:

\begin{equation}
    \rm m_{ij}' = \phi_e\left(\rm h_i^l, \rm h_j^l, \| Q\rm x_i^l + g - Q\rm x_j^l - g \|^2, e_{ij} \right) = \rm m_{ij}
\end{equation}

The position update in ESSGNN (adapted from EGNN) is defined as:

\begin{equation}
    \rm x_i^{l+1} = \rm x_i^l + \sum_{j \ne i} (\rm x_i^l - \rm x_j^l) \cdot \phi_x(\rm m_{ij})
\end{equation}

We now show that this equation is SE(3) equivariant. Applying the transformation:

\begin{align*}
    Q\rm x_i^l + g + \sum_{j \ne i} \left( Q\rm x_i^l + g - Q\rm x_j^l - g \right) \cdot \phi_x(\rm m_{ij}) &= Q\rm x_i^l + g + Q \sum_{j \ne i} (\rm x_i^l - \rm x_j^l) \cdot \phi_x(\rm m_{ij}) \\
    &= Q \left( \rm x_i^l + \sum_{j \ne i} (\rm x_i^l - \rm x_j^l) \cdot \phi_x(\rm m_{ij}) \right) + g \\
    &= Q\rm x_i^{l+1} + g
\end{align*}

Thus, the coordinate update is SE(3) equivariant.

For the feature update:
\begin{equation}
    \rm h_i^{l+1} = \rm h_i^l + \sum_{j \ne i} \phi_h(\rm m_{ij})
\end{equation}

Since \( \rm m_{ij} \) is invariant to transformations of \( \rm x \), and both \( \rm h_i^l \), \( \rm h_j^l \) and \( e_{ij} \) are independent of the global pose, the feature update is invariant to SE(3) transformations of positions.

Therefore, the ESSGNN update satisfies:

\begin{equation}
    Q\rm x^{l+1} + g,\; \rm h^{l+1} = \mathrm{ESSGNN}(Q\rm x^l + g,\; \rm h^l,\; E)
\end{equation}

This completes the proof that ESSGNN preserves SE(3) equivariance despite the inclusion of semantic edge embeddings. 

\clearpage

\section{Experimental Analysis} \label{sec:experimental analysis}

\begin{figure*}[h]
  \centering

  \begin{subfigure}[b]{0.45\textwidth}
    \centering
    \includegraphics[width=\linewidth]{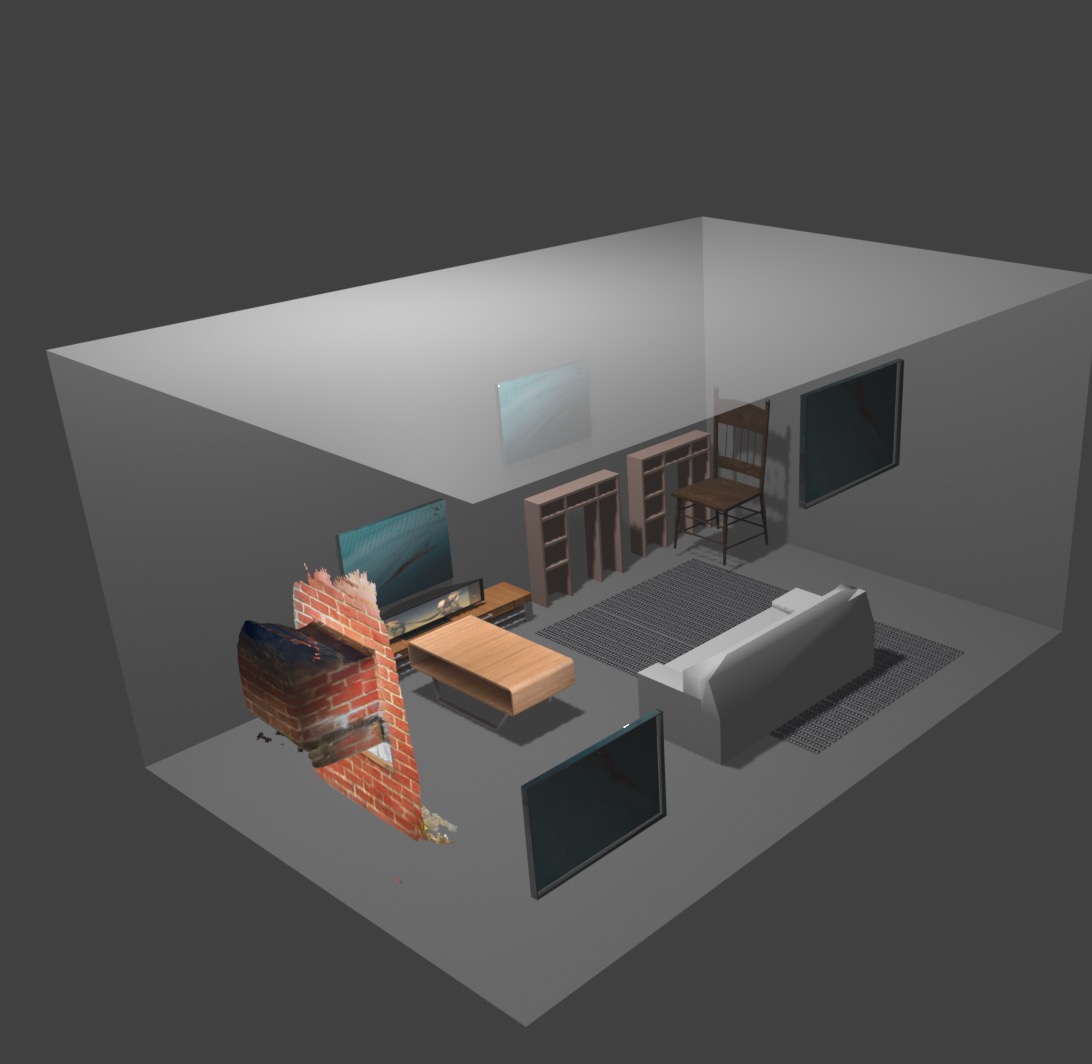}
    \caption{Without ESSGNN encoder}
  \end{subfigure}
  \hfill
  \begin{subfigure}[b]{0.45\textwidth}
    \centering
    \includegraphics[width=\linewidth]{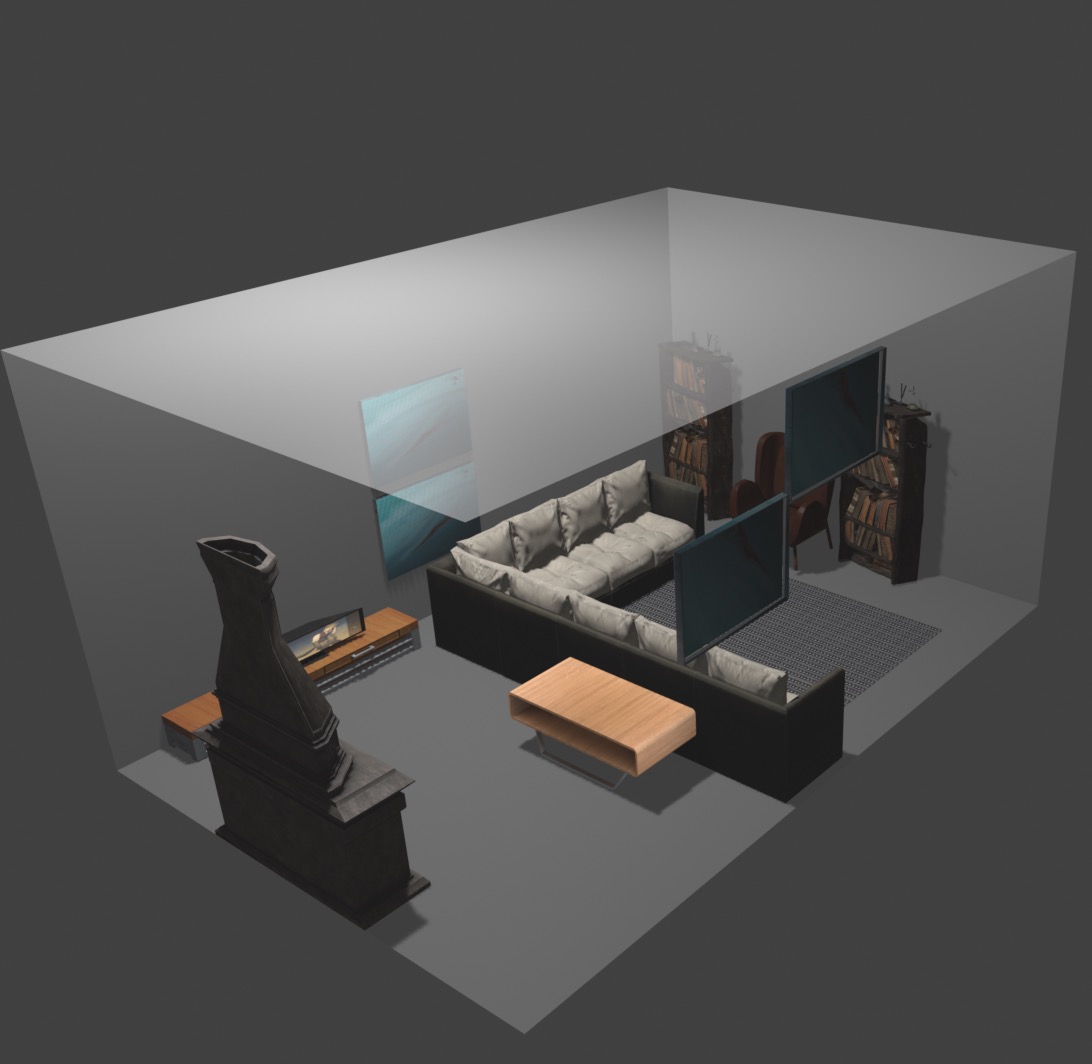}
    \caption{With ESSGNN encoder}
  \end{subfigure}

  \vspace{2mm}
  \caption*{\textbf{Room 1 Description:} A classical-style lounge for group leisure and conversation}

  \vspace{5mm}

  \begin{subfigure}[b]{0.45\textwidth}
    \centering
    \includegraphics[width=\linewidth]{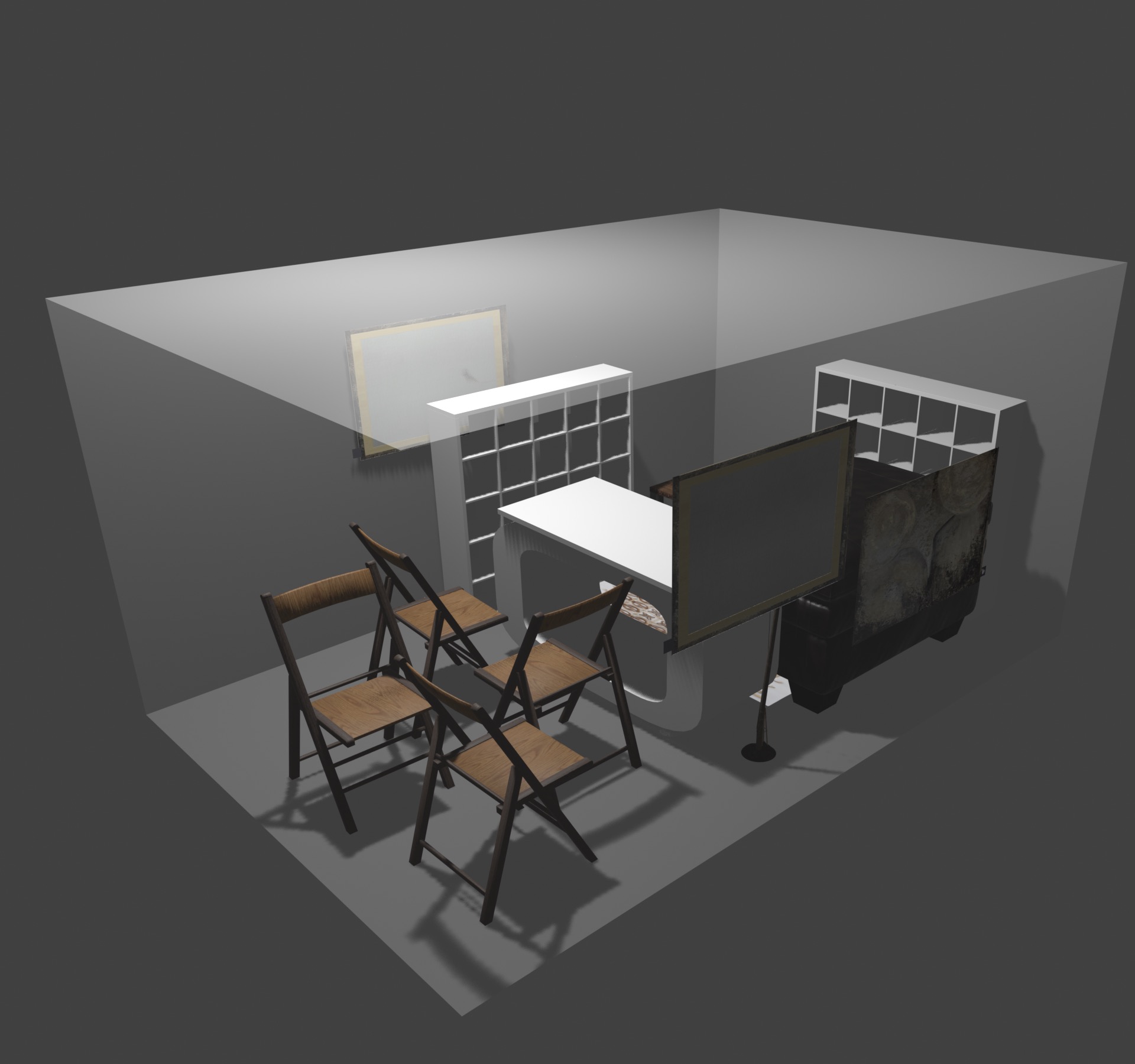}
    \caption{Without ESSGNN encoder}
  \end{subfigure}
  \hfill
  \begin{subfigure}[b]{0.45\textwidth}
    \centering
    \includegraphics[width=\linewidth]{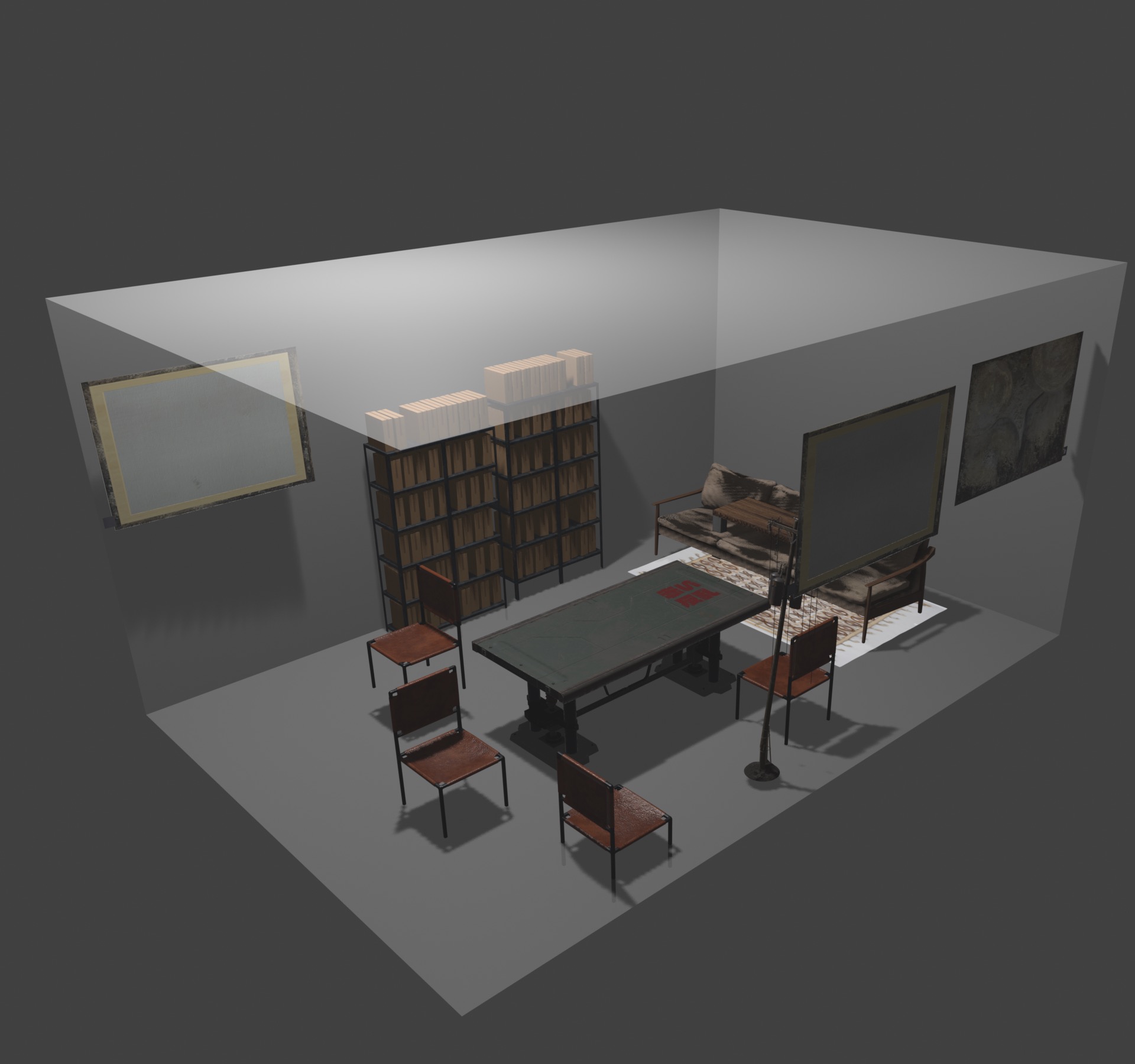}
    \caption{With ESSGNN encoder}
  \end{subfigure}

  \vspace{2mm}
  \caption*{\textbf{Room 2 Description:} An aged archive room for research and consultation}

  \caption{Comparison of scene generation with and without ESSGNN encoder.}
  \label{fig:essgnn_comparison}
\end{figure*}

As shown in Figure~\ref{fig:essgnn_comparison}:

\textbf{Room 1}: Without ESSGNN, the room lacks stylistic coherence—the metallic fireplace and mismatched furniture deviate from the classical theme. With ESSGNN, the scene adopts a unified classical aesthetic with a dark-toned fireplace, matching sofa, and bookshelf.

\textbf{Room 2}: Without ESSGNN, modern office furniture and cluttered seating break the archive theme and hinder functionality. With ESSGNN, compact wooden chairs are arranged around the table, better fitting the aged archive context and improving usability.

%% file: 4backgound.tex
\section{Related Work}
3D scene generation serves as the broader task context of our work, encompassing both generative and retrieval-based approaches to assembling realistic virtual environments. Within this paradigm, 3D object retrieval plays a critical role by providing high-quality assets that satisfy semantic, stylistic, and spatial constraints. We first review recent advances in scene generation frameworks, followed by an overview of representative models for multimodal 3D object retrieval.

\subsection{3D Scene Generation Paradigms}

Recent progress in 3D scene generation follows two directions. The first relies on generative models that synthesize entire 3D scenes in mesh, voxel, or neural field formats~\cite{schult2024controlroom3d}. While promising, these methods struggle with ensuring object-level realism or semantic fidelity~\cite{fang2023ctrl}. To address these limitations, a second paradigm emerges that frames scene generation as a layout composition task using retrieved assets from large-scale 3D repositories. LLMs and VLMs exhibit advanced capabilities in various tasks \cite{pan2025evomarlcoevolutionarymultiagentreinforcement,pan2025advevomarlshapinginternalizedsafety,pan2025fairreasonbalancingreasoningsocial}: software engineering \citep{pan2025code,pan2024codevbenchllmsunderstanddevelopercentric}, question answering systems \citep{pan2024chain,pan2024convcoaimprovingopendomainquestion}, and scientific discovery \citep{shao2025sciscigpt}. Methods like LayoutGPT~\cite{feng2023layoutgpt} and I-Design~\cite{ccelen2024design} employ LLMs as planners to generate layouts from text descriptions. More recent techniques, such as LayoutVLM~\cite{sun2024layoutvlm}, improve physical plausibility through differentiable rendering optimization and layout supervision from image-marked datasets. Despite their advances, they still face two fundamental challenges: (1) limited internalized 3D spatial reasoning within VLMs and (2) the inefficiency and poor generalization of supervised fine-tuning, which relies on scarce and imperfect layout annotations. MetaSpatial~\cite{pan2025metaspatial} addresses these issues via a reinforcement learning-based framework that optimizes 3D spatial layouts in real time using physics-aware constraints and rendered-image evaluations. This significantly enhances scene plausibility and coherence.

While MetaSpatial focuses on improving reasoning in layout generation, another crucial but underexplored dimension is the design of the retrieval mechanism itself. Most prior works rely on general-purpose models, such as OpenShape~\cite{liu2023openshape}, to fetch 3D assets. However, these models are not specifically trained for multimodal, scene-conditioned retrieval. They struggle to support arbitrary combinations of user inputs (e.g., missing modality scenarios) and treat object retrieval as an independent parallel process, neglecting layout dependencies. To bridge this gap, we propose a retrieval-centric framework that explicitly incorporates layout context into the retrieval loop. Unlike prior work, our method supports arbitrary modality combinations, performs iterative context-aware retrieval, and introduces a plug-and-play ESSGNN module to encode scene layout as a structured graph. This enables spatially consistent and stylistically coherent scene construction.

\subsection{3D Object Retrieval}
3D object retrieval has traditionally focused on aligning visual and geometric representations of objects with semantic queries in the form of text, image, or point cloud inputs. Early approaches rely on contrastive learning between 2D/3D pairs, such as PointCLIP~\cite{zhang2022pointclip} and CLIP-Forge~\cite{sanghi2022clip}, which repurpose vision-language models for shape retrieval. More recent methods like ULIP~\cite{xue2024ulip} and OpenShape~\cite{liu2023openshape} extend this to tri-modal alignment, embedding text, image, and 3D point clouds into a unified latent space via either single-tower or dual-tower architectures. However, these models are trained purely on object-centric data and assume complete modality availability, limiting their robustness under missing or partial query inputs. Beyond alignment, retrieval models such as SCA3D~\cite{ren2025sca3d} and COM3D~\cite{wu2024com3d} improve representation quality via self-augmentation or compositional reasoning, yet still lack explicit mechanisms to handle arbitrary modality combinations or incorporate contextual cues. OmniBind~\cite{wang2024omnibindlargescaleomnimultimodal} offers more flexible modality binding but is not optimized for retrieval tasks involving spatial constraints. In contrast, MetaFind is explicitly designed for context-aware, multimodal 3D asset retrieval. Our model supports free-form modality combinations and is robust to missing inputs through stochastic masking. Most notably, it augments retrieval with scene context by incorporating an ESSGNN-based layout encoder, enabling iterative, layout-aware asset selection that better supports spatial realism and scene consistency.